\documentclass[11pt]{article}
\usepackage{acl2015}
\usepackage{times}
\usepackage{url}
\usepackage{latexsym}
\usepackage{natbib}
\usepackage{graphics}
\usepackage{amsmath}
\usepackage{amssymb}
\usepackage{booktabs}
\usepackage{graphicx}
\usepackage{tikz}
\usetikzlibrary{positioning,bayesnet}

\title{Improved Semantic Representations From \\ Tree-Structured Long Short-Term Memory Networks}

\author{Kai Sheng Tai, Richard Socher*, Christopher D. Manning \\
  Computer Science Department, Stanford University, *MetaMind Inc. \\
  {\tt kst@cs.stanford.edu, richard@metamind.io, manning@stanford.edu}}

\date{}

\begin{document}
\maketitle
\begin{abstract}
Because of their superior ability to preserve sequence information over time, Long Short-Term Memory (LSTM) networks, a type of recurrent neural network with a more complex computational unit, have obtained strong results on a variety of sequence modeling tasks. The only underlying LSTM structure that has been explored so far is a linear chain. However, natural language exhibits syntactic properties that would naturally combine words to phrases.
We introduce the Tree-LSTM, a generalization of LSTMs to tree-structured network topologies. Tree-LSTMs outperform all existing systems and strong LSTM baselines on two tasks: predicting the semantic relatedness of two sentences (SemEval 2014, Task 1) and sentiment classification (Stanford Sentiment Treebank).

\end{abstract}

\section{Introduction}		

Most models for distributed representations of phrases and sentences---that is, models where real-valued vectors are used to represent meaning---fall into one of three classes: bag-of-words models, 
 sequence models, and tree-structured models. In bag-of-words models, phrase and sentence representations are independent of word order; for example, they can be generated by averaging constituent word representations \citep{landauer1997solution,foltz1998measurement}.
  In contrast, sequence models construct sentence representations as an order-sensitive function of the sequence of tokens \citep{elman1990finding,mikolov2012statistical}.
   Lastly, tree-structured models compose each phrase and sentence representation from its constituent subphrases according to a given syntactic structure over the sentence \citep{goller1996learning,socher2011parsing}.

Order-insensitive models are insufficient to fully capture the semantics of natural language due to their inability to account for differences in meaning as a result of differences in word order or syntactic structure (\emph{e.g.,} ``cats climb trees'' \emph{vs.} ``trees climb cats'').  We therefore turn to order-sensitive sequential or tree-structured models. 
 In particular, tree-structured models are a linguistically attractive option due to their relation to syntactic interpretations of sentence structure.   A natural question, then, is the following: to what extent (if at all) can we do better with tree-structured models as opposed to sequential models for sentence representation? In this paper, we work towards addressing this question by directly comparing a type of sequential model that has recently been used to achieve state-of-the-art results in several NLP tasks against its tree-structured generalization.

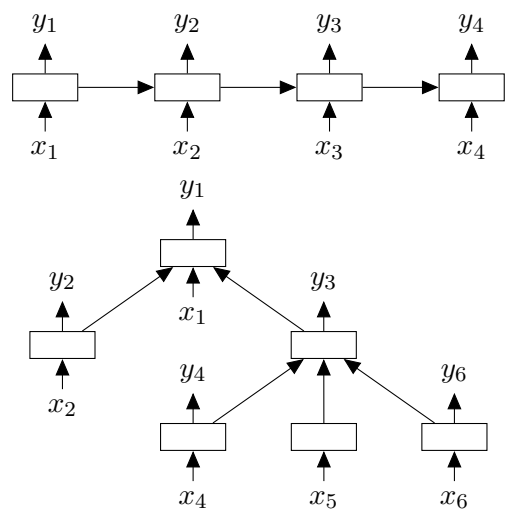
\begin{figure}[t]
\begin{center}
\begin{tikzpicture}
  \node[draw,text width=6mm,text height=1mm] (1) {};
  \node[draw,text width=6mm,text height=1mm, right=of 1] (2) {};
  \node[draw,text width=6mm,text height=1mm, right=of 2] (3) {};
  \node[draw,text width=6mm,text height=1mm, right=of 3] (4) {};
  \node[below=4mm of 1] (x1) {$x_1$};
  \node[below=4mm of 2] (x2) {$x_2$};
  \node[below=4mm of 3] (x3) {$x_3$};
  \node[below=4mm of 4] (x4) {$x_4$};
  \node[above=4mm of 1] (y1) {$y_1$};
  \node[above=4mm of 2] (y2) {$y_2$};
  \node[above=4mm of 3] (y3) {$y_3$};
  \node[above=4mm of 4] (y4) {$y_4$};
  
  \edge {1} {2};
  \edge {2} {3};
  \edge {3} {4};
  \edge {x1} {1};
  \edge {x2} {2};
  \edge {x3} {3};
  \edge {x4} {4};
  \edge {1} {y1};
  \edge {2} {y2};
  \edge {3} {y3};
  \edge {4} {y4};
\end{tikzpicture}

\begin{tikzpicture}
  \node[draw,text width=6mm,text height=1mm] (1) {};
  \node[draw,text width=6mm,text height=1mm, below left=12mm of 1] (2) {};
  \node[draw,text width=6mm,text height=1mm, below right=12mm of 1] (3) {};
  \node[draw,text width=6mm,text height=1mm, below left=12mm of 3] (4) {};
  \node[draw,text width=6mm,text height=1mm, right=8.5mm of 4](5) {};
  \node[draw,text width=6mm,text height=1mm, below right=12mm of 3] (6) {};
  \node[below=4mm of 1] (x1) {$x_1$};
  \node[below=4mm of 2] (x2) {$x_2$};
  \node[below=4mm of 4] (x4) {$x_4$};
  \node[below=4mm of 5] (x5) {$x_5$};
  \node[below=4mm of 6] (x6) {$x_6$};
  \node[above=4mm of 1] (y1) {$y_1$};
  \node[above=4mm of 2] (y2) {$y_2$};
  \node[above=4mm of 3] (y3) {$y_3$};
  \node[above=4mm of 4] (y4) {$y_4$};
  \node[above=4mm of 6] (y6) {$y_6$};
  
  \edge {2,3,x1} {1};
  \edge {4,5,6} {3};
  \edge {x2} {2};
  \edge {x4} {4};
  \edge {x5} {5};
  \edge {x6} {6};
  \edge {1} {y1};
  \edge {2} {y2};
  \edge {3} {y3};
  \edge {4} {y4};
  \edge {6} {y6};
\end{tikzpicture}
\end{center}
\caption{\textbf{Top:} A chain-structured LSTM network. \textbf{Bottom:} A tree-structured LSTM network with arbitrary branching factor.
}
\end{figure}

Due to their capability for processing arbitrary-length sequences, recurrent neural networks (RNNs) are a natural choice for sequence modeling tasks. Recently, RNNs with Long Short-Term Memory (LSTM) units \citep{hochreiter1997long} have re-emerged as a popular architecture due to their representational power and effectiveness at capturing long-term dependencies. LSTM networks, which we review in Sec.~\ref{sec:lstms}, have been successfully applied to a variety of sequence modeling and prediction tasks, notably machine translation \citep{bahdanau2014neural,sutskever2014sequence}, speech recognition \citep{graves2013hybrid}, image caption generation \citep{vinyals2014show}, and program execution \citep{zaremba2014learning}.

In this paper, we introduce a generalization of the standard LSTM architecture to tree-structured network topologies and show its superiority for representing sentence meaning over a sequential LSTM. While the standard LSTM composes its hidden state from the input at the current time step and the hidden state of the LSTM unit in the previous time step, the tree-structured LSTM, or Tree-LSTM, composes its state from an input vector and the hidden states of arbitrarily many child units. The standard LSTM can then be considered a special case of the Tree-LSTM where each internal node has exactly one child.

In our evaluations, we demonstrate the empirical strength of Tree-LSTMs as models for representing sentences. We evaluate the Tree-LSTM architecture on two tasks: semantic relatedness prediction on sentence pairs and sentiment classification of sentences drawn from movie reviews. Our experiments show that Tree-LSTMs outperform existing systems and sequential LSTM baselines on both tasks. Implementations of our models and experiments are available at \url{https://github.com/stanfordnlp/treelstm}.

\section{Long Short-Term Memory Networks}
\label{sec:lstms}

\subsection{Overview}

Recurrent neural networks (RNNs) are able to process input sequences of arbitrary length via the recursive application of a transition function on a \emph{hidden state vector} $h_t$. At each time step~$t$, the hidden state~$h_t$ is a function of the input vector~$x_t$ that the network receives at time $t$ and its previous hidden state~$h_{t-1}$. For example, the input vector~$x_t$ could be a vector representation of the $t$-th word in body of text \citep{elman1990finding,mikolov2012statistical}. The hidden state~$h_t \in \mathbb{R}^d$ can be interpreted as a $d$-dimensional distributed representation of the sequence of tokens observed up to time~$t$.

Commonly, the RNN transition function is an affine transformation followed by a pointwise nonlinearity such as the hyperbolic tangent function:
\begin{equation*}
h_t = \tanh\left( Wx_t + Uh_{t-1} + b \right).
\end{equation*}
Unfortunately, a problem with RNNs with transition functions of this form is that during training, components of the gradient vector can grow or decay exponentially over long sequences \citep{hochreiter1998vanishing,bengio1994learning}.  
This problem with \emph{exploding} or \emph{vanishing gradients} makes it difficult for the RNN model to learn long-distance correlations in a sequence.

The LSTM architecture \citep{hochreiter1997long} addresses this problem of learning long-term dependencies by introducing a \emph{memory cell} that is able to preserve state over long periods of time. While numerous LSTM variants have been described, here we describe the version used by \citet{zaremba2014learning}.

We define the LSTM \emph{unit} at each time step~$t$ to be a collection of vectors in $\mathbb{R}^d$: an \emph{input gate}~$i_t$, a \emph{forget gate}~$f_t$,  an \emph{output gate}~$o_t$, a \emph{memory cell}~$c_t$ and a hidden state~$h_t$. The entries of the gating vectors $i_t$, $f_t$ and $o_t$ are in $[0, 1]$. We refer to $d$ as the \emph{memory dimension} of the LSTM.

The LSTM transition equations are the following:
\begin{align}
i_t &= \sigma\left( W^{(i)} x_t + U^{(i)} h_{t-1} + b^{(i)} \right), \label{eq:lstm-eqs} \\
f_t &= \sigma\left( W^{(f)} x_t + U^{(f)} h_{t-1} + b^{(f)} \right), \nonumber \\
o_t &= \sigma\left( W^{(o)} x_t + U^{(o)} h_{t-1} + b^{(o)} \right), \nonumber \\
u_t &= \tanh \left( W^{(u)} x_t + U^{(u)} h_{t-1} + b^{(u)} \right), \nonumber \\
c_t &=  i_t \odot u_t + f_t \odot c_{t-1}, \nonumber \\
h_t &= o_t \odot \tanh(c_t), \nonumber
\end{align}
where $x_t$ is the input at the current time step, $\sigma$ denotes the logistic sigmoid function and $\odot$ denotes elementwise multiplication. Intuitively, the forget gate controls the extent to which the previous memory cell is forgotten, the input gate controls how much each unit is updated, and the output gate controls the exposure of the internal memory state. The hidden state vector in an LSTM unit is therefore a gated, partial view of the state of the unit's internal memory cell. Since the value of the gating variables vary for each vector element, the model can learn to represent information over multiple time scales.

\subsection{Variants}

Two commonly-used variants of the basic LSTM architecture are the Bidirectional LSTM and the Multilayer LSTM (also known as the \emph{stacked} or \emph{deep} LSTM).

\paragraph{Bidirectional LSTM.} A Bidirectional LSTM \citep{graves2013hybrid} consists of two LSTMs that are run in parallel: one on the input sequence and the other on the reverse of the input sequence. At each time step, the hidden state of the Bidirectional LSTM is the concatenation of the forward and backward hidden states. This setup allows the hidden state to capture both past and future information.

\paragraph{Multilayer LSTM.} In Multilayer LSTM architectures, the hidden state of an LSTM unit in layer $\ell$ is used as input to the LSTM unit in layer $\ell + 1$ in the same time step \citep{graves2013hybrid,sutskever2014sequence,zaremba2014learning}. Here, the idea is to let the higher layers capture longer-term dependencies of the input sequence.

\smallskip
These two variants can be combined as a Multilayer Bidirectional LSTM \citep{graves2013hybrid}.

\section{Tree-Structured LSTMs}
\label{sec:tree-structured-lstms}

A limitation of the LSTM architectures described in the previous section is that they only allow for strictly sequential information propagation. Here, we propose two natural extensions to the basic LSTM architecture: the \emph{Child-Sum Tree-LSTM} and the \emph{N-ary Tree-LSTM}. Both variants allow for richer network topologies where each LSTM unit is able to incorporate information from multiple child units.

As in standard LSTM units, each Tree-LSTM unit (indexed by $j$)  contains input and output gates~$i_j$ and $o_j$, a memory cell~$c_j$ and hidden state~$h_j$. The difference between the standard LSTM unit and Tree-LSTM units is that gating vectors and memory cell updates are dependent on the states of possibly many child units. Additionally, instead of a single forget gate, the Tree-LSTM unit contains one forget gate~$f_{jk}$ for each child~$k$. This allows the Tree-LSTM unit to selectively incorporate information from each child. For example, a Tree-LSTM model can learn to emphasize semantic heads in a semantic relatedness task, or it can learn to preserve the representation of sentiment-rich children for sentiment classification.

As with the standard LSTM, each Tree-LSTM unit takes an input vector $x_j$. In our applications, each $x_j$ is a vector representation of a word in a sentence. The input word at each node depends on the tree structure used for the network. For instance, in a Tree-LSTM over a dependency tree, each node in the tree takes the vector corresponding to the head word as input, whereas in a Tree-LSTM over a constituency tree, the leaf nodes take the corresponding word vectors as input.

\begin{figure}[t]
\begin{center}
\begin{tikzpicture}

\node[latent] (h1) {$h_1$};
\node[latent, left=of h1] (c1) {$c_1$};
\node[latent, left=of c1] (u1) {$u_1$};
\node[latent, left=of u1] (x1) {$x_1$};
\node[latent, below=7mm of u1] (c3) {$c_3$};
\node[latent, above=7mm of u1] (c2) {$c_2$};
\node[latent, below=7mm of x1] (h3) {$h_3$};
\node[latent, above=7mm of x1] (h2) {$h_2$};

\edge {x1,h2,h3} {u1};
\draw[->] (c2) to node[above right=-1mm] {$f_2$} (c1);
\draw[->] (c3) to node[below right=-1mm] {$f_3$} (c1);
\draw[->] (u1) to node[above=-0.5mm] {$i_1$} (c1);
\draw[->] (c1) to node[above=-0.5mm] {$o_1$} (h1);

\node[draw,fit=(h2) (c2)] {};
\node[draw,fit=(h3) (c3)] {};
\node[draw,fit=(c1) (h1)] {};

\end{tikzpicture}
\end{center}
\caption{Composing the memory cell $c_1$ and hidden state $h_1$ of a Tree-LSTM unit with two children (subscripts 2 and 3). Labeled edges correspond to gating by the indicated gating vector, with dependencies omitted for compactness.}
\end{figure}
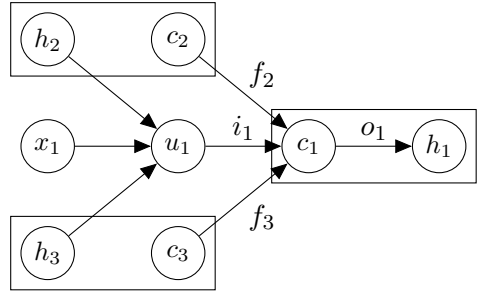

\subsection{Child-Sum Tree-LSTMs}
\label{sec:child-sum-treelstms}

Given a tree, let $C(j)$ denote the set of children of node $j$. The Child-Sum Tree-LSTM transition equations are the following:
\begin{align}
\tilde{h}_j &= \sum_{k \in C(j)} h_k, \label{eq:treelstm-first} \\
i_j &=\sigma \left( W^{(i)} x_j + U^{(i)} \tilde{h}_j + b^{(i)} \right), \\
f_{jk} &= \sigma\left( W^{(f)} x_j + U^{(f)} h_k + b^{(f)} \right), \label{eq:treelstm-f}\\
o_j &= \sigma \left( W^{(o)} x_j + U^{(o)} \tilde{h}_j  + b^{(o)} \right), \\
u_j &= \tanh\left( W^{(u)} x_j + U^{(u)} \tilde{h}_j  + b^{(u)} \right), \\
c_j &= i_j \odot u_j + \sum_{k\in C(j)} f_{jk} \odot c_{k}, \\
h_j &= o_j \odot \tanh(c_j), \label{eq:treelstm-last}
\end{align}
where in Eq.~\ref{eq:treelstm-f}, $k \in C(j)$.

Intuitively, we can interpret each parameter matrix in these equations as encoding correlations between the component vectors of the Tree-LSTM unit, the input $x_j$, and the hidden states $h_k$ of the unit's children. For example, in a dependency tree application, the model can learn parameters $W^{(i)}$ such that the components of the input gate $i_j$ have values close to 1 (\emph{i.e.}, ``open'') when a semantically important content word (such as a verb) is given as input, and values close to 0 (\emph{i.e.}, ``closed'') when the input is a relatively unimportant word (such as a determiner).

\paragraph{Dependency Tree-LSTMs.} Since the Child-Sum Tree-LSTM unit conditions its components on the sum of child hidden states $h_k$, it is well-suited for trees with high branching factor or whose children are unordered. For example, it is a good choice for dependency trees, where the number of dependents of a head can be highly variable. We refer to a Child-Sum Tree-LSTM applied to a dependency tree as a \mbox{\emph{Dependency Tree-LSTM}}.

\subsection{$N$-ary Tree-LSTMs}
\label{sec:nary-treelstms}

The $N$-ary Tree-LSTM can be used on tree structures where the branching factor is at most $N$ and where children are ordered, \emph{i.e.}, they can be indexed from $1$ to $N$. For any node $j$, write the hidden state and memory cell of its $k$th child as $h_{jk}$ and $c_{jk}$ respectively. The $N$-ary Tree-LSTM transition equations are the following:
\begin{align}
i_j &=\sigma \left( W^{(i)} x_j + \sum_{\ell=1}^N U^{(i)}_\ell h_{j\ell} + b^{(i)} \right), \label{eq:nary-treelstm-first}\\
f_{jk} &= \sigma\left( W^{(f)} x_j + \sum_{\ell=1}^N U^{(f)}_{k\ell} h_{j\ell} + b^{(f)} \right), \label{eq:nary-treelstm-f}\\
o_j &= \sigma \left( W^{(o)} x_j + \sum_{\ell=1}^N U^{(o)}_\ell h_{j\ell}  + b^{(o)} \right), \\
u_j &= \tanh\left( W^{(u)} x_j + \sum_{\ell=1}^N U^{(u)}_\ell h_{j\ell}  + b^{(u)} \right), \\
c_j &= i_j \odot u_j + \sum_{\ell=1}^N f_{j\ell} \odot c_{j\ell}, \\
h_j &= o_j \odot \tanh(c_j), \label{eq:nary-treelstm-last}
\end{align}
where in Eq.~\ref{eq:nary-treelstm-f}, $k = 1, 2, \dots, N$. Note that when the tree is simply a chain, both Eqs.~\ref{eq:treelstm-first}--\ref{eq:treelstm-last} and Eqs.~\ref{eq:nary-treelstm-first}--\ref{eq:nary-treelstm-last} reduce to the standard LSTM transitions, Eqs.~\ref{eq:lstm-eqs}.

The introduction of separate parameter matrices for each child $k$ allows the $N$-ary Tree-LSTM model to learn more fine-grained conditioning on the states of a unit's children than the Child-Sum Tree-LSTM. Consider, for example, a constituency tree application where the left child of a node corresponds to a noun phrase, and the right child to a verb phrase. Suppose that in this case it is advantageous to emphasize the verb phrase in the representation. Then the $U^{(f)}_{k\ell}$ parameters can be trained such that the components of $f_{j1}$ are close to 0 (\emph{i.e.}, ``forget''), while the components of $f_{j2}$ are close to 1 (\emph{i.e.}, ``preserve'').

\paragraph{Forget gate parameterization.} In Eq.~\ref{eq:nary-treelstm-f}, we define a parameterization of the $k$th child's forget gate~$f_{jk}$ that contains ``off-diagonal'' parameter matrices $U^{(f)}_{k\ell}$, $k \neq \ell$. This parameterization allows for more flexible control of information propagation from child to parent. For example, this allows the left hidden state in a binary tree to have either an \emph{excitatory} or \emph{inhibitory} effect on the forget gate of the right child. However, for large values of $N$, these additional parameters are impractical and may be tied or fixed to zero.

\paragraph{Constituency Tree-LSTMs.} We can naturally apply \emph{Binary} Tree-LSTM units to binarized constituency trees since left and right child nodes are distinguished. We refer to this application of Binary Tree-LSTMs as a \mbox{\emph{Constituency Tree-LSTM}}. Note that in Constituency Tree-LSTMs, a node $j$ receives an input vector~$x_j$ only if it is a leaf node.

\medskip
In the remainder of this paper, we focus on the special cases of Dependency Tree-LSTMs and Constituency Tree-LSTMs. These architectures are in fact closely related; since we consider only binarized constituency trees, the parameterizations of the two models are very similar. The key difference is in the application of the compositional parameters: dependent \emph{vs.} head for Dependency Tree-LSTMs, and left child \emph{vs.} right child for Constituency Tree-LSTMs.

\section{Models}

We now describe two specific models that apply the Tree-LSTM architectures described in the previous section.

\subsection{Tree-LSTM Classification}
\label{sec:classification-model}

In this setting, we wish to predict labels $\hat{y}$ from a discrete set of classes $\mathcal{Y}$ for some subset of nodes in a tree. For example, the label for a node in a parse tree could correspond to some property of the phrase spanned by that node.

At each node $j$, we use a softmax classifier to predict the label $\hat{y}_j$ given the inputs $\{x\}_j$ observed at nodes in the subtree rooted at $j$. The classifier takes the hidden state $h_j$ at the node as input: 
\begin{align*}
\hat{p}_\theta(y ~|~ \{ x\}_j) &= \mathrm{softmax}\left( W^{(s)} h_j + b^{(s)} \right), \\ 
\hat{y_j} &= \arg\max_{y} \hat{p}_\theta\left(y ~|~ \{x\}_j\right).
\end{align*}

The cost function is the negative log-likelihood of the true class labels $y^{(k)}$ at each labeled node:
\begin{equation*}
J(\theta) = -\frac{1}{m} \sum_{k=1}^m \log \hat{p}_\theta \Big(y^{(k)} ~\Big|~ \{x\}^{(k)}\Big) + \frac{\lambda}{2} \| \theta \|_2^2,
\end{equation*}
where $m$ is the number of labeled nodes in the training set, the superscript $k$ indicates the $k$th labeled node, and $\lambda$ is an L2 regularization hyperparameter.

\subsection{Semantic Relatedness of Sentence Pairs}
\label{sec:sim-model}

Given a sentence pair, we wish to predict a real-valued similarity score in some range $[1, K]$, where $K > 1$ is an integer. The sequence $\{1, 2, \dots, K\}$ is some ordinal scale of similarity, where higher scores indicate greater degrees of similarity, and we allow real-valued scores to account for ground-truth ratings that are an average over the evaluations of several human annotators.

We first produce sentence representations $h_L$ and $h_R$ for each sentence in the pair using a Tree-LSTM model over each sentence's parse tree. Given these sentence representations, we predict the similarity score $\hat{y}$ using a neural network that considers both the distance and angle between the pair $(h_L, h_R)$:
\begin{align}
h_\times &= h_L \odot h_R, \label{eq:sim-network} \\
h_+ &= |h_L - h_R|, \nonumber \\
h_s &= \sigma\left(W^{(\times)} h_\times  + W^{(+)} h_+ + b^{(h)} \right), \nonumber \\
\hat{p}_\theta     &= \mathrm{softmax}\left(W^{(p)} h_s + b^{(p)} \right), \nonumber \\
\hat{y}     &= r^T \hat{p}_\theta, \nonumber
\end{align}
where $r^T = [1~2~\dots~K]$ and the absolute value function is applied elementwise. The use of both distance measures $h_\times$ and $h_+$ is empirically motivated: we find that the combination outperforms the use of either measure alone. The  multiplicative measure~$h_\times$ can be interpreted as an elementwise comparison of the signs of the input representations.

We want the expected rating under the predicted distribution $\hat{p}_\theta$ given model parameters $\theta$ to be close to the gold rating $y \in [1, K]$: $\hat{y} = r^T \hat{p}_\theta \approx y$. We therefore define a sparse target distribution\footnote{In the subsequent experiments, we found that optimizing this objective yielded better performance than a mean squared error objective.} $p$ that satisfies $y = r^T p$:
\begin{equation*}
p_i = \begin{cases} y - \lfloor y \rfloor, & i = \lfloor y \rfloor + 1 \\ \lfloor y \rfloor - y + 1, & i = \lfloor y \rfloor  \\ 0 & \text{otherwise}\end{cases}
\end{equation*}
for $1 \leq i \leq K$. The cost function is the regularized KL-divergence between $p$ and $\hat{p}_\theta$:
\begin{equation*}
 J(\theta) = \frac{1}{m} \sum_{k=1}^m \mathrm{KL}\Big(p^{(k)}~\Big\|~\hat{p}^{(k)}_\theta\Big) +  \frac{\lambda}{2} \| \theta \|_2^2,
\end{equation*}
where $m$ is the number of training pairs and the superscript $k$ indicates the $k$th sentence pair.

\section{Experiments}

We evaluate our Tree-LSTM architectures on two tasks: (1) sentiment classification of sentences sampled from movie reviews and (2) predicting the semantic relatedness of sentence pairs.

In comparing our Tree-LSTMs against sequential LSTMs, we control for the number of LSTM parameters by varying the dimensionality of the hidden states\footnote{For our Bidirectional LSTMs, the parameters of the forward and backward transition functions are shared. In our experiments, this achieved superior performance to Bidirectional LSTMs with untied weights and the same number of parameters (and therefore smaller hidden vector dimensionality).}. Details for each model variant are summarized in Table~\ref{tab:parameter-counts}.

\subsection{Sentiment Classification}

In this task, we predict the sentiment of sentences sampled from movie reviews. We use the Stanford Sentiment Treebank \citep{socher2013recursive}. There are two subtasks: binary classification of sentences, and fine-grained classification over five classes: very negative, negative, neutral, positive, and very positive. We use the standard train/dev/test splits of 6920/872/1821 for the binary classification subtask and 8544/1101/2210 for the fine-grained classification subtask (there are fewer examples for the binary subtask since neutral sentences are excluded). Standard binarized constituency parse trees are provided for each sentence in the dataset, and each node in these trees is annotated with a sentiment label.

For the sequential LSTM baselines, we predict the sentiment of a phrase using the representation given by the final LSTM hidden state. The sequential LSTM models are trained on the spans corresponding to labeled nodes in the training set.

We use the classification model described in Sec.~\ref{sec:classification-model} with both Dependency Tree-LSTMs (Sec.~\ref{sec:child-sum-treelstms}) and Constituency Tree-LSTMs (Sec.~\ref{sec:nary-treelstms}). The Constituency Tree-LSTMs are structured according to the provided parse trees. For the Dependency Tree-LSTMs, we produce dependency parses\footnote{Dependency parses produced by the Stanford Neural Network Dependency Parser \citep{chen2014fast}.} of each sentence; each node in a tree is given a sentiment label if its span matches a labeled span in the training set. 

\begin{table}[t]
\resizebox{\columnwidth}{!}{
\begin{tabular}{rcccc} \toprule
& \multicolumn{2}{c}{\textbf{Relatedness}} & \multicolumn{2}{c}{\textbf{Sentiment}} \\ \cmidrule{2-5}
\textbf{LSTM Variant} & $d$ & $|\theta|$ & $d$ & $|\theta|$ \ \\ \midrule
Standard & 150 & 203,400 & 168 & 315,840 \\ 
Bidirectional & 150 & 203,400 & 168 & 315,840 \\ 
2-layer & 108 & 203,472 & 120 & 318,720 \\
Bidirectional 2-layer & 108 & 203,472 & 120 & 318,720 \\
Constituency Tree & 142 & 205,190  & 150 & 316,800 \\
Dependency Tree & 150 & 203,400 & 168 & 315,840 \\ \bottomrule
\end{tabular}
}
\caption{Memory dimensions $d$ and composition function parameter counts $|\theta|$ for each LSTM variant that we evaluate.}
\label{tab:parameter-counts}
\end{table}

\subsection{Semantic Relatedness}

For a given pair of sentences, the semantic relatedness task is to predict a human-generated rating of the similarity of the two sentences in meaning.

We use the Sentences Involving Compositional Knowledge (SICK) dataset \citep{marelli2014semeval}, consisting of 9927 sentence pairs in a 4500/500/4927 train/dev/test split. The sentences are derived from existing image and video description datasets. Each sentence pair is annotated with a relatedness score $y \in [1,5]$, with 1 indicating that the two sentences are completely unrelated, and 5 indicating that the two sentences are very related. Each label is the average of 10 ratings assigned by different human annotators.

Here, we use the similarity model described in Sec.~\ref{sec:sim-model}. For the similarity prediction network (Eqs.~\ref{eq:sim-network}) we use a hidden layer of size 50. We produce binarized constituency parses\footnote{Constituency parses produced by the Stanford PCFG Parser \citep{klein2003accurate}.} and dependency parses of the sentences in the dataset for our Constituency Tree-LSTM and Dependency Tree-LSTM models.

\begin{table}[t]
\begin{center}
\resizebox{\columnwidth}{!}{
\begin{tabular}{lcc}
\toprule \bf Method & \bf Fine-grained & \bf Binary \\
\midrule
RAE \citep{socher2013recursive} & 43.2 & 82.4 \\ 
MV-RNN \citep{socher2013recursive} & 44.4 & 82.9 \\
RNTN \citep{socher2013recursive} & 45.7 & 85.4 \\
DCNN \citep{blunsom2014convolutional} & 48.5 & 86.8 \\
Paragraph-Vec \citep{le2014distributed}  & 48.7 & 87.8 \\ 
CNN-non-static \citep{kim2014convolutional} & 48.0 & 87.2 \\
CNN-multichannel \citep{kim2014convolutional} & 47.4 & \textbf{88.1} \\
DRNN \citep{irsoy2014deep} & 49.8 & 86.6 \\
\midrule
LSTM                                	 & 46.4 \:(1.1) & 84.9 \:(0.6) \\
Bidirectional LSTM   		 & 49.1 \:(1.0) & 87.5 \:(0.5) \\
2-layer LSTM 			 & 46.0 \:(1.3) & 86.3 \:(0.6) \\
2-layer Bidirectional LSTM & 48.5 \:(1.0) & 87.2 \:(1.0) \\ \midrule
Dependency Tree-LSTM 	 & 48.4 \:(0.4) & 85.7 \:(0.4) \\
Constituency Tree-LSTM 	 & & \\
\quad -- randomly initialized vectors & 43.9 \:(0.6) & 82.0 \:(0.5)\\
\quad -- Glove vectors, fixed & 49.7 \:(0.4) & 87.5 \:(0.8) \\
\quad -- Glove vectors, tuned &  \textbf{51.0} \:(0.5) & 88.0 \:(0.3)\\ \bottomrule
\end{tabular}
}
\end{center}
\caption{Test set accuracies on the Stanford Sentiment Treebank. For our experiments, we report mean accuracies over 5 runs (standard deviations in parentheses). \textbf{Fine-grained:} 5-class sentiment classification. \textbf{Binary:} positive/negative sentiment classification.}
\label{tab:sentiment-results}
\end{table}

\begin{table*}[t]
\begin{center}
\resizebox{0.85\textwidth}{!}{
\begin{tabular}{lccc}
\toprule \bf Method & \bf Pearson's $r$ & \bf Spearman's $\rho$ & \bf MSE \\ \midrule
Illinois-LH \citep{lai2014illinois} & 0.7993  & 0.7538 & 0.3692 \\
UNAL-NLP \citep{jimenez2014unal} & 0.8070 & 0.7489 & 0.3550 \\
Meaning Factory \citep{bjerva2014meaning}  & 0.8268 & 0.7721 & 0.3224 \\
ECNU \citep{zhao2014ecnu} & 0.8414 & -- & -- \\ \midrule
Mean vectors 				& 0.7577 \enspace(0.0013) & 0.6738 \enspace(0.0027) & 0.4557 \enspace(0.0090) \\
DT-RNN \citep{socher2014grounded} & 0.7923 \enspace(0.0070) & 0.7319 \enspace(0.0071) & 0.3822 \enspace(0.0137)  \\
SDT-RNN \citep{socher2014grounded}  & 0.7900 \enspace(0.0042) & 0.7304 \enspace(0.0076) &  0.3848 \enspace(0.0074) \\ \midrule
LSTM					& 0.8528 \enspace(0.0031)  & 0.7911 \enspace(0.0059) & 0.2831 \enspace(0.0092)  \\
Bidirectional LSTM 			& 0.8567 \enspace(0.0028)  & 0.7966 \enspace(0.0053) & 0.2736 \enspace(0.0063) \\
2-layer LSTM 				& 0.8515 \enspace(0.0066)  & 0.7896 \enspace(0.0088) & 0.2838 \enspace(0.0150) \\ 
2-layer Bidirectional LSTM 	& 0.8558 \enspace(0.0014)  & 0.7965 \enspace(0.0018) & 0.2762 \enspace(0.0020) \\ \midrule
Constituency Tree-LSTM 		& 0.8582 \enspace(0.0038)  & 0.7966 \enspace(0.0053) & 0.2734 \enspace(0.0108)  \\
Dependency Tree-LSTM 		& \textbf{0.8676} \enspace(0.0030) & \textbf{0.8083} \enspace(0.0042) & \textbf{0.2532} \enspace(0.0052) \\ \bottomrule
\end{tabular}
}
\end{center}
\caption{Test set results on the SICK semantic relatedness subtask. For our experiments, we report mean scores over 5 runs (standard deviations in parentheses). Results are grouped as follows: \textbf{(1)} SemEval 2014 submissions; \textbf{(2)} Our own baselines; \textbf{(3)} Sequential LSTMs; \textbf{(4)} Tree-structured LSTMs.}
\label{tab:semanticsim-results}
\end{table*}

\subsection{Hyperparameters and Training Details}

The hyperparameters for our models were tuned on the development set for each task.

We initialized our word representations using publicly available 300-dimensional Glove vectors\footnote{Trained on 840 billion tokens of Common Crawl data, \url{http://nlp.stanford.edu/projects/glove/}.} \citep{pennington2014glove}. For the sentiment classification task, word representations were updated during training with a learning rate of 0.1. For the semantic relatedness task, word representations were held fixed as we did not observe any significant improvement when the representations were tuned.

Our models were trained using AdaGrad \citep{duchi2011adaptive} with a learning rate of 0.05 and a minibatch size of 25. The model parameters were regularized with a per-minibatch L2 regularization strength of $10^{-4}$. The sentiment classifier was additionally regularized using dropout \citep{hinton2012improving} with a dropout rate of 0.5. We did not observe performance gains using dropout on the semantic relatedness task.

\section{Results}
\label{sec:results}

\subsection{Sentiment Classification}

Our results are summarized in Table~\ref{tab:sentiment-results}. The Constituency Tree-LSTM outperforms existing systems on the fine-grained classification subtask and achieves accuracy comparable to the state-of-the-art on the binary subtask. In particular, we find that it outperforms the Dependency Tree-LSTM.  This performance gap is at least partially attributable to the fact that the Dependency Tree-LSTM is trained on less data: about 150K labeled nodes \emph{vs.} 319K for the Constituency Tree-LSTM. This difference is due to (1) the dependency representations containing fewer nodes than the corresponding constituency representations, and (2) the inability to match about 9\% of the dependency nodes to a corresponding span in the training data.

We found that updating the word representations during training (``fine-tuning'' the word embedding) yields a significant boost in performance on the fine-grained classification subtask and gives a minor gain on the binary classification subtask (this finding is consistent with previous work on this task by \citet{kim2014convolutional}). These gains are to be expected since the Glove vectors used to initialize our word representations were not originally trained to capture sentiment.

\subsection{Semantic Relatedness}

Our results are summarized in Table~\ref{tab:semanticsim-results}. Following \citet{marelli2014semeval}, we use Pearson's $r$, Spearman's $\rho$ and mean squared error (MSE) as evaluation metrics. The first two metrics are measures of correlation against human evaluations of semantic relatedness.

We compare our models against a number of non-LSTM baselines. The mean vector baseline computes sentence representations as a mean of the representations of the constituent words. The DT-RNN and SDT-RNN models \citep{socher2014grounded} both compose vector representations for the nodes in a dependency tree as a sum over affine-transformed child vectors, followed by a nonlinearity. The SDT-RNN is an extension of the DT-RNN that uses a separate transformation for each dependency relation. For each of our baselines, including the LSTM models, we use the similarity model described in Sec.~\ref{sec:sim-model}.

We also compare against four of the top-performing systems\footnote{We list the strongest results we were able to find for this task; in some cases, these results are stronger than the official performance by the team on the shared task. For example, the listed result by \citet{zhao2014ecnu} is stronger than their submitted system's Pearson correlation score of 0.8280.} submitted to the SemEval 2014 semantic relatedness shared task: ECNU \citep{zhao2014ecnu}, The Meaning Factory \citep{bjerva2014meaning}, UNAL-NLP  \citep{jimenez2014unal}, and Illinois-LH \citep{lai2014illinois}. These systems are heavily feature engineered, generally using a combination of surface form overlap features and lexical distance features derived from WordNet or the Paraphrase Database \citep{ganitkevitch2013ppdb}.

Our LSTM models outperform all these systems without any additional feature engineering, with the best results achieved by the Dependency Tree-LSTM. Recall that in this task, both Tree-LSTM models only receive supervision at the root of the tree, in contrast to the sentiment classification task where supervision was also provided at the intermediate nodes. We conjecture that in this setting, the Dependency Tree-LSTM benefits from its more compact structure relative to the Constituency Tree-LSTM, in the sense that paths from input word vectors to the root of the tree are shorter on aggregate for the Dependency Tree-LSTM.

\begin{table*}[t]
\begin{center}
\resizebox{\textwidth}{!}{
\begin{tabular}{cc}
    \begin{tabular}{lc}
    \toprule \bf Ranking by mean word vector cosine similarity  & \bf Score \\ \midrule
     \bf{a woman is slicing potatoes} & \\
    a woman is cutting potatoes & 0.96 \\
    a woman is slicing herbs & 0.92\\
    a woman is slicing tofu & 0.92\\  \midrule
    \bf{a boy is waving at some young runners from the ocean} & \\
    a man and a boy are standing at the bottom of some stairs , & 0.92\\ \quad  which are outdoors &  \\
    a group of children in uniforms is standing at a gate and & 0.90\\ \quad one is kissing the mother &  \\
    a group of children in uniforms is standing at a gate and  & 0.90\\ \quad there is no one kissing the mother &  \\ \midrule
    \bf{two men are playing guitar} & \\
    some men are playing rugby & 0.88 \\
    two men are talking & 0.87 \\
     & \\
    two dogs are playing with each other & 0.87 \\ \bottomrule
    \end{tabular} &
    
    \begin{tabular}{lc}
    \toprule \bf Ranking by Dependency Tree-LSTM model & \bf Score \\ \midrule
    \bf{a woman is slicing potatoes} & \\
    a woman is cutting potatoes & 4.82\\
    potatoes are being sliced by a woman & 4.70\\
    tofu is being sliced by a woman & 4.39\\ \midrule
    \bf{a boy is waving at some young runners from the ocean} & \\
    a group of men is playing with a ball on the beach & 3.79 \\
    & \\
    a young boy wearing a red swimsuit is jumping out of a & 3.37 \\ \quad blue kiddies pool &  \\
    the man is tossing a kid into the swimming pool that is  & 3.19\\ \quad near the ocean &  \\ \midrule
    \bf{two men are playing guitar} & \\
    the man is singing and playing the guitar & 4.08 \\
    the man is opening the guitar for donations and plays & 4.01 \\ \quad with the case & \\
    two men are dancing and singing in front of a crowd & 4.00 \\ \bottomrule
    \end{tabular} \\
\end{tabular}
}
\caption{Most similar sentences from a 1000-sentence sample drawn from the SICK test set. The Tree-LSTM model is able to pick up on more subtle relationships, such as that between ``beach'' and ``ocean'' in the second example.}
\label{tab:sim-examples}
\end{center}
\end{table*}

\section{Discussion and Qualitative Analysis}
\label{sec:discussion}

\begin{figure}[t]
\centering
\includegraphics[width=\columnwidth]{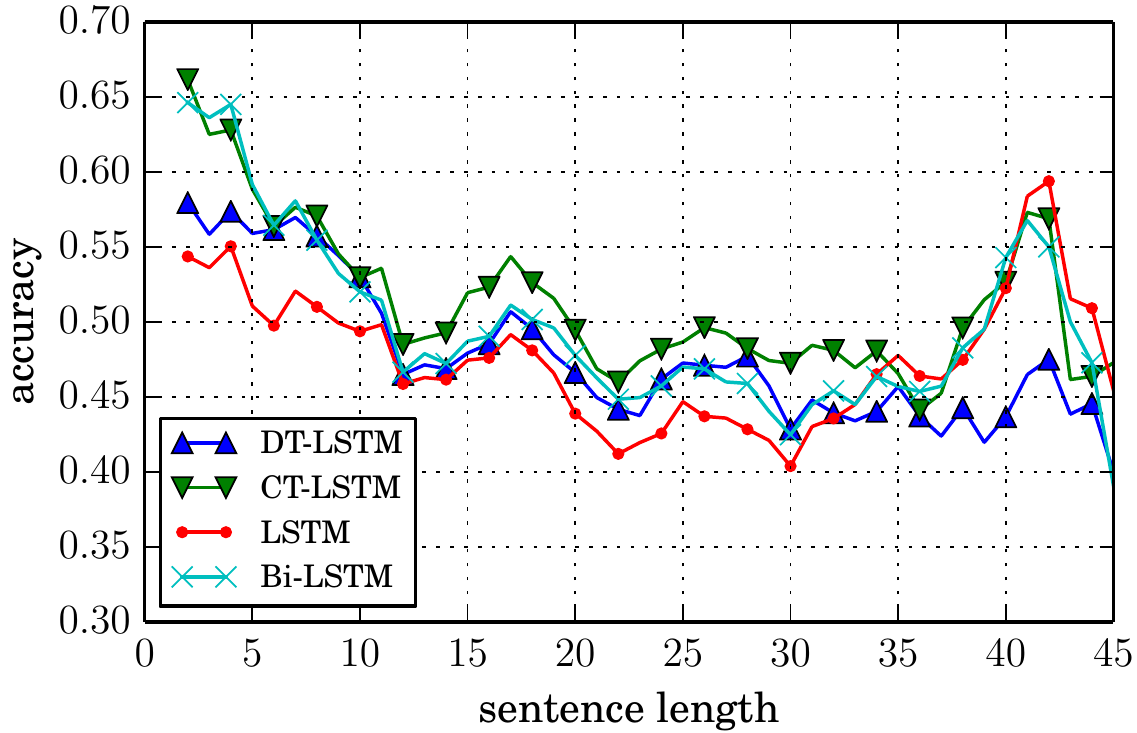}
\caption{Fine-grained sentiment classification accuracy \emph{vs.} sentence length. For each $\ell$, we plot accuracy for the test set sentences with length in the window $[\ell - 2, \ell + 2]$. Examples in the tail of the length distribution are batched in the final window ($\ell = 45$).}
\label{fig:sent-acc-length}
\end{figure}

\subsection{Modeling Semantic Relatedness}

In Table~\ref{tab:sim-examples}, we list nearest-neighbor sentences retrieved from a 1000-sentence sample of the SICK test set. We compare the neighbors ranked by the Dependency Tree-LSTM model against a baseline ranking by cosine similarity of the mean word vectors for each sentence.

The Dependency Tree-LSTM model exhibits several desirable properties. Note that in the dependency parse of the second query sentence, the word ``ocean'' is the second-furthest word from the root (``waving''), with a depth of 4. Regardless, the retrieved sentences are all semantically related to the word ``ocean'', which indicates that the Tree-LSTM is able to both preserve and emphasize information from relatively distant nodes. Additionally, the Tree-LSTM model shows greater robustness to differences in sentence length. Given the query ``two men are playing guitar'', the Tree-LSTM associates the phrase ``playing guitar'' with the longer, related phrase ``dancing and singing in front of a crowd'' (note as well that there is zero token overlap between the two phrases).

\subsection{Effect of Sentence Length}

\begin{figure}[t]
\centering
\includegraphics[width=\columnwidth]{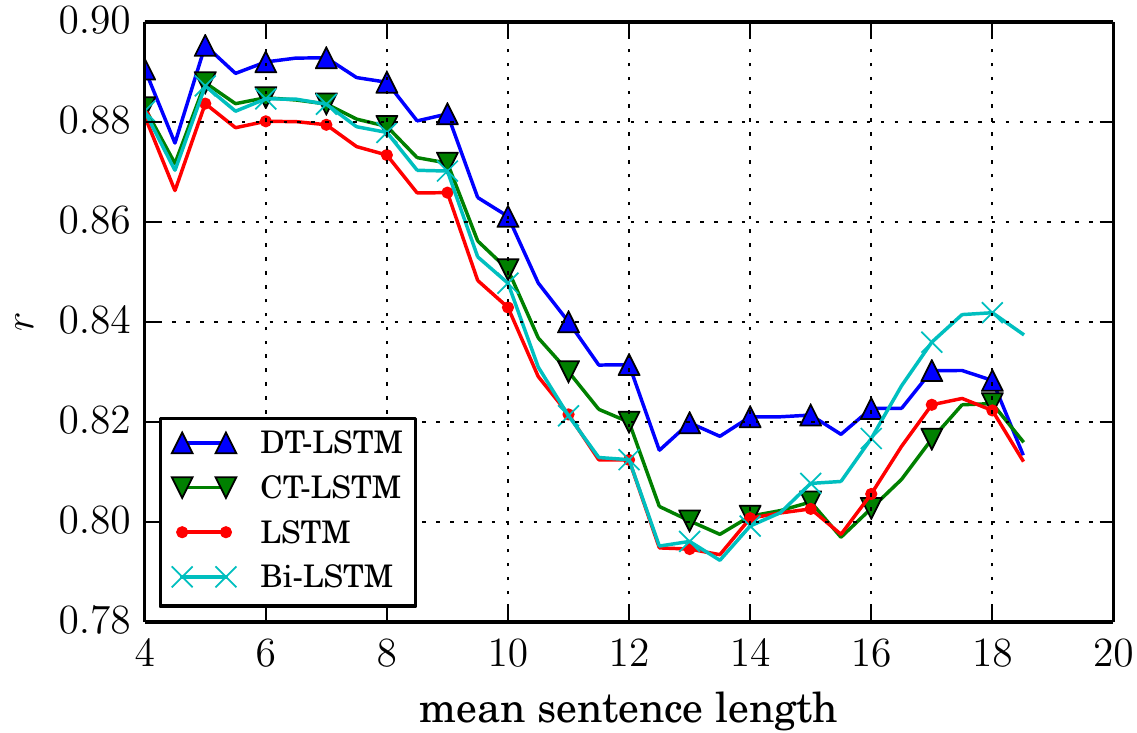}
\caption{Pearson correlations $r$ between predicted similarities and gold ratings \emph{vs.} sentence length. For each $\ell$, we plot $r$ for the pairs with mean length in the window $[\ell - 2, \ell + 2]$. Examples in the tail of the length distribution are batched in the final window ($\ell = 18.5$).}
\label{fig:sim-pearson-length}
\end{figure}

One hypothesis to explain the empirical strength of Tree-LSTMs is that tree structures help mitigate the problem of preserving state over long sequences of words. If this were true, we would expect to see the greatest improvement over sequential LSTMs on longer sentences. In Figs.~\ref{fig:sent-acc-length} and \ref{fig:sim-pearson-length}, we show the relationship between sentence length and performance as measured by the relevant task-specific metric. Each data point is a mean score over 5 runs, and error bars have been omitted for clarity.

We observe that while the Dependency Tree-LSTM does significantly outperform its sequential counterparts on the relatedness task for longer sentences of length 13 to 15 (Fig.~\ref{fig:sim-pearson-length}), it also achieves consistently strong performance on shorter sentences. This suggests that unlike sequential LSTMs, Tree-LSTMs are able to encode semantically-useful structural information in the sentence representations that they compose.

\section{Related Work}

Distributed representations of words  \citep{rumelhart1988learning,collobert2011natural,turian2010word,HuangEtAl2012,mikolov2013distributed,pennington2014glove} have found wide applicability in a variety of NLP tasks. Following this success, there has been substantial interest in the area of learning distributed phrase and sentence representations \citep{mitchell2010composition,yessenalina2011compositional,grefenstette2013multi,mikolov2013distributed}, as well as distributed representations of longer bodies of text such as paragraphs and documents \citep{srivastava2013modeling,le2014distributed}.

Our approach builds on recursive neural networks \citep{goller1996learning,socher2011parsing}, which we abbreviate as Tree-RNNs in order to avoid confusion with recurrent neural networks. Under the Tree-RNN framework, the vector representation associated with each node of a tree is composed as a function of the vectors corresponding to the children of the node. The choice of composition function gives rise to numerous variants of this basic framework. Tree-RNNs have been used to parse images of natural scenes \citep{socher2011parsing}, compose phrase representations from word vectors \citep{socher2012semantic}, and classify the sentiment polarity of sentences \citep{socher2013recursive}. 

\section{Conclusion}

In this paper, we introduced a generalization of LSTMs to tree-structured network topologies. The Tree-LSTM architecture can be applied to trees with arbitrary branching factor. We demonstrated the effectiveness of the Tree-LSTM by applying the architecture in two tasks: semantic relatedness and sentiment classification, outperforming existing systems on both. Controlling for model dimensionality, we demonstrated that Tree-LSTM models are able to outperform their sequential counterparts. Our results suggest further lines of work in characterizing the role of structure in producing distributed representations of sentences. 

\section*{Acknowledgements}

We thank our anonymous reviewers for their valuable feedback. 
Stanford University gratefully acknowledges the support of a Natural Language
Understanding-focused gift from Google Inc. and the Defense
Advanced Research Projects Agency (DARPA) Deep Exploration and Filtering
of Text (DEFT) Program under Air Force Research Laboratory (AFRL)
contract no. FA8750-13-2-0040. Any opinions, findings, and conclusion or
recommendations expressed in this material are those of the authors and
do not necessarily reflect the view of the DARPA, AFRL, or the US
government.

\bibliographystyle{acl}
\bibliography{treelstm}

\end{document}